\documentclass[journal]{IEEEtran}
\usepackage{cite}
\usepackage{lipsum}

\ifCLASSINFOpdf
\else
\fi

\usepackage{graphicx}
\graphicspath{{image/}}
\usepackage{url}
\usepackage{graphicx}
\usepackage{color}
\usepackage{amsmath}
\usepackage{amssymb}
\usepackage{subfigure}
\usepackage{soul}

\begin{document}
\title{Efficient Video Summarization Framework using EEG and Eye-tracking Signals}

\author{Sai~Sukruth~Bezugam*,~\IEEEmembership{Student Member,~IEEE,}
        Swatilekha~Majumdar*, Chetan~Ralekar*
        and~Tapan~Kumar~Gandhi$^+$,~\IEEEmembership{Senior~Member,~IEEE}% <-this % stops a space
\IEEEcompsocitemizethanks{
%\IEEEcompsocthanksitem 
 %S. S. Bezugam is affiliated to the Department of Electrical Engineering, Indian Institute of Technology, Delhi, India.
%E-mail : saisukruthbezugam@ieee.org
% \IEEEcompsocthanksitem 
% S. Majumdar is affiliated to the Department of Electrical Engineering, Indian Institute of Technology, Delhi, India.
%E-mail : swatilekha.phd@gmail.com
% \IEEEcompsocthanksitem 
% C. Ralekar is affiliated to the Department of Electrical Engineering, Indian Institute of Technology, Delhi, India.
%E-mail : chetan.ralekar@gmail.com 
\IEEEcompsocthanksitem 
S. S. Bezugam, S. Majumdar, C. Ralekar, T. K. Gandhi is affiliated to the Department of Electrical Engineering, Indian Institute of Technology, Delhi, India.
 E-mail : {saisukruthbezugam@ieee.org, swatilekha.phd@gmail.com, chetan.ralekar@gmail.com, tgandhi@iitd.ac.in} \protect\\
 \hfil\
\hfil\
%\IEEEcompsocthanksitem J. Doe and J. Doe are with Anonymous University.
}% <-this % stops an unwanted space
\thanks{*Authors have contributed equally to the paper.}
\thanks{$^+$(Corresponding Author: T. K. Gandhi).}
\thanks{Manuscript received XXXX, XXXX; revised XXXX.}}


\markboth{This manuscript is submitted for review}%
{Bezugam \MakeLowercase{\textit{et al.}}:Efficient Video Summarization Framework using EEG and Eye Tracking Signals}



\maketitle

\begin{abstract}
This paper proposes an efficient video summarization framework that will give a gist of the entire video in a few key-frames or video skims. Existing video summarization frameworks are based on algorithms that utilize computer vision low-level feature extraction or high-level domain level extraction. However, being the ultimate user of the summarized video, humans remain the most neglected aspect. Therefore, the proposed paper considers human's role in summarization and introduces human visual attention-based summarization techniques. To understand human attention behavior, we have designed and performed experiments with human participants using electroencephalogram (EEG) and eye-tracking technology. The EEG and eye-tracking data obtained from the experimentation are processed simultaneously and used to segment frames containing useful information from a considerable video volume. Thus, the frame segmentation primarily relies on the cognitive judgments of human beings. Using our approach, a video is summarized by $\sim$96.5\% while maintaining higher precision ($\sim$0.98) and high recall factors ($\sim$0.97). The comparison with the state-of-the-art techniques demonstrates that the proposed approach yields ceiling-level performance with reduced computational cost in summarising the videos.
\end{abstract}


% \fi
\IEEEpeerreviewmaketitle

\section{Introduction}
\IEEEPARstart{R}{ecent} advances in storage and digital media technology have made recording and aggregating massive volumes of videos facile \cite{xu2016heterogeneous}. An enormous amount of videos are uploaded to video-sharing websites every minute. Moreover, a large volume of the surveillance videos is recorded to ensure security, and analyzing such videos is tedious. It is very challenging to retrieve relevant information from vast video content in an efficient manner. Addressing these hurdles, efforts are being made to devise a video summary, which gives the essence of the entire video in a short time. This technique is known as video summarization \cite{nasir2019fog,lai2016video,zhang2016context}. Video summarization is a method that expedites faster browsing of extensive video collections and facilitates dynamic content indexing and access. Thus, video summarization is getting more importance in the multimedia industry \cite{potapov2014category,mundur2006keyframe,ma2002user,ji2019video}, especially in the field of surveillance. This summarization is achieved by an efficient selection of key-frames that are representative of the whole video. The extraction of key-frames is performed by detecting change points, low-level features based on clustering, or clustering depending on objects. The key-frames are helpful for indexing videos; however, they are void of motion data. This, in turn, restricts their use in information retrieval tasks. In video skimming, short segments of the video are selected for summarization. Such segments should be properly chosen in order to convey maximum information. 

It is very well known that humans are the end-users of these summarization tasks who want to get the gist of an hourly long video in a couple of minutes. Hence, it is of utmost importance that inputs are taken from humans for designing a video summarization framework \cite{mehmood2016divide,wang2014hybrid}. However, the state of the art deep learning-based methods hardly consider humans' role in proposing such architectures. It is imperative to consider how humans perceive the video and extract critical information. Thus, the present paper attempts to extract the key segments from the video by understanding perceptual processes and visual attention. This paper takes into account human perception while extracting key-frames.

The major contributions of the paper can be summarized as:
\begin{itemize}
    \item We proposed a video summarization technique using cues from perceptual and visual attentive behavior of humans. 
    \item We have designed, and performed experiments using both EEG and eye-tracking technology and used the respective signals to extract the key-frames.
    \item We have compared the proposed technique with state-of-the-art video summarization techniques.
\end{itemize}

The paper's organization is as follows: Section II highlights some related works in video summarization that have been proposed in the literature. The experimental details and setup description for the proposed methodology has been discussed in section III. The proposed video summarization framework is described in section IV. Section V discusses the performance of the proposed framework and establishes performance benchmarks compared to some recent works proposed in the literature. The section also elaborates on the advantage of using both EEG and eye-tracking in video summarization. Section V gives concluding remarks with possible future extensions.

\section{Related Works}
Video summarization is a technique which aids faster browsing of extensive video collections leading to more efficient content indexing and access. The summary is generated by extracting key-frames that best portray the video. While selecting the segments, care should be taken to embody the whole video and segregate important information to the user. Traditional video summarization techniques can be roughly classified into three categories: (i) Summarization based on hand-crafted features, (ii) Summarization frameworks based on deep learning architecture, (iii) Summarization based on human perception.

\subsection{Summarization based on on hand-crafted features}
A large part of video summarization research has focused on the development of two major techniques $-$ (i) low-level video summarization techniques, where videos are summarised by examining the low-level features such as the shape, color, motion, and speech present in the video \cite{money2010elvis}, and (ii) high-level or domain-specific video summarization, where domain-specific information is used to recognize quintessential shots of a video. %In literature, Ejaz et al. \cite{ejaz2012adaptive}, Zhou et al. \cite{zhou2010feature}, De et al. \cite{de2011vsumm}, Furini et al. \cite{furini2010stimo}, and Almeida et al. \cite{almeida2012vison} have used low-level features to summarize video rushes. 
Ejaz et al. \cite{ejaz2012adaptive} used an aggregation mechanism to combine visual features, where correlations between RGB color channels, color histogram, and moments of inertia lead to extraction of keyframes from a video sequence. Similarly, Zhou et al. \cite{zhou2010feature} extracted audio, color, and motion features, which were fused using an adaptive weighting mechanism. The video sequence is then clustered by using a fuzzy c-means scheme along with an optimally determined cluster number. A summarization technique based on color feature extraction was presented by De et al. \cite{de2011vsumm}. %In this method, the frames are grouped in a sequential order instead of being randomly distributed between clusters and are then grouped using the traditional k-means algorithm, leading to the generation of a summary selecting one frame per cluster. 
Similarly, a summarization system for producing on-the-fly video storyboards was presented in \cite{furini2010stimo}. This method creates still and moving storyboards with additional features of advanced customization by users. 
%This method is based on a fast clustering algorithm, which selects the most descriptive visual frames based on the Hue-Saturation Value (HSV) frame color distribution. For each frame in the input sequence, the visual features are extracted to describe the visual content. After extracting the features, a fast and straightforward algorithm is used to detect groups of video frames with similar content and to select a representative frame from each group. 
A segmentation-based clustering algorithm was presented by Almeida et al. in \cite{almeida2012vison}. In this case, the video was broken down into a set of essential shots by decomposing the image stream's color histogram. Next, in each video shot, a Zero-mean Normalized Cross-correlation metric \cite{martin1995comparison} is employed to eliminate redundant frames. The chosen frames are then refined to bypass any insignificant frames in the video summary. Joho et al. \cite{joho2009exploiting} followed a summarization process based on the viewer's facial expression. A video summarization based on pupillary dilation and eye gaze was proposed by Katti et al. \cite{katti2011affective}. Mehmood et al., in their paper \cite{mehmood2016divide}, used EEG signals obtained from visual, auditory, and data from beta ($\beta$) frequency band to extract keyframes from the video. 
%The methodology utilizes a shot-boundary detection based divide-and-conquer approach, the viewer's neuronal responses as a potential source of information, and an efficient intra- and inter-modality attention fusion for video summarization. 

From the above description, we can see that while low-level video summarization methods mostly rely on the contents of the frame to summarize the video. Recently, some papers have also used EEG and fMRI signals to do the same. Eliciting the informative contents using the high-frequency components of EEG signals, Salehin et al. \cite{salehin2017affective} proposed a method for video summarization. In this, the Empirical Mode Decomposition (EMD) technique was employed to disintegrate EEG signals into high to low-frequency components known as Intrinsic Mode Function (IMF). Video frames tagged and summarized as per emotions, as recognized from EEG was proposed in \cite{singhal2018summarization}. The authors further used a crowd-sourcing model to develop the condition and evaluation of video summarization. Han et al. \cite{han2014video} employed fMRI signals from the brain to guide the extraction of visually informative segments from videos. These techniques relied on handcrafted features, which may or may not fetch the best summary from the video. Therefore, deep learning-based approaches were introduced.

\subsection{Summarization frameworks based on on deep learning architecture}
AI-based summarization tackles the problem of summarizing a video mostly based on clustering techniques. These solutions have a two-step approach: (a) distinguishing video shots from the video, and (b) choosing key-frames based on some criterion from each video shot. Zhuang et al. \cite{zhuang1998adaptive} and Hanjalic et al. \cite{hanjalic1997new} used color histogram features in the HSV color space to proposed an unsupervised clustering approach. In \cite{hanjalic1997new}, the key-frames are chosen by considering the frame closest to the cluster centroid. Gong and Liu \cite{gong2001video} proposed a video summarization where the input video to be summarized was clustered to form sets. Although it was expected that the key-frames in the same cluster should be visually the same, the sets obtained in this case were visually different from each other.
All the above methods act on a rule-based approach and hardly consider how humans perceive things. The deep learning based video summarization frameworks like VSUMM \cite{de2008vsumm} uses highly complex architectures which are hardly explainable. Thus, these frameworks are used as black boxes. Recently, explainable architectures \cite{selvaraju2017grad},\cite{chattopadhay2018grad} are proposed to understand the black box. However, such architecture generates heat maps, which highlights the important image regions; humans remains the most neglected aspect in those architectures.
%XAI refers to methods and techniques in the application of artificial intelligence technology (AI) such that the results of the solution can be understood by human experts. It contrasts with the concept of the ``black box'' in machine learning where even their designers cannot explain why the AI arrived at a specific decision.

\subsection{Summarization based on human perception}
Most of the video summarization methodologies that are presented in literature either focus on visual, auditory, or facial expression data of the user. However, some researchers have started using EEG signals along with the above methodologies to summarize the video to obtain useful frames. Mehmood et al. \cite{mehmood2015audio} proposed a summing scheme that combined EEG signals with audio-visual features for comprehension of videos. Salehin and Paul \cite{salehin2017affective} introduced an amalgamation of Empirical Mode Decomposition (EMD) and EEG to summarize video events. In EMD, different frequency components (high to low) was derived from EEG signals to obtain key-frames from the digital video. EEG seemed to be a choice for some researchers \cite{salehin2017affective,kim2019video,kim2010toward,kim2017video} while extracting key-frames for video summarization. However, the proposed framework was driven by `how humans perceive what they see and observe'. This can not be addressed without using the eye-tracking technique because eye-movements can provide moment-to-moment information processing \cite{rayner1998eye} by the brain. Moreover, it is assumed that humans fixate on the object of interest to them. Thus, the eye-tracking technology has been widely used to identify essential image regions for the task of image recognition \cite{ralekar2017unlocking,ralekar2019intelligent}, memorability detection \cite{khanna2019memorability}, along with understanding the perceptual process. Therefore along with the EEG, it is vital to consider the eye-tracking events to get an overall picture of which key-frames are necessary for us. 

%\IEEEpubidadjcol
\section{Experimental Details}
\begin{figure}[t]
    \centering
    \includegraphics[width=0.25\textwidth]{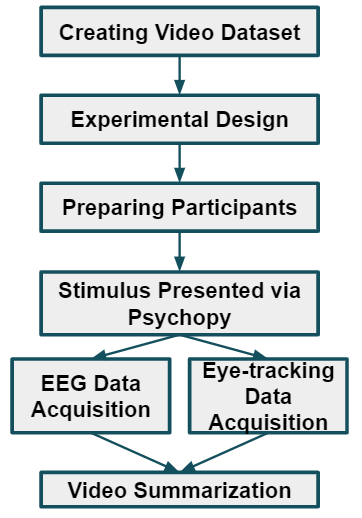}
    \caption{Methodology followed for data acquisition and pre-processing the the work presented in the paper.}
    \label{f4}
\end{figure}

The overall experimental details are given in Fig. \ref{f4}. The methodology involves creating a video dataset, experimental design, participant screening, stimulus preparation for data acquisition, acquiring data.
\begin{figure}[t]
    \centering
    {\includegraphics[width=0.48\textwidth]{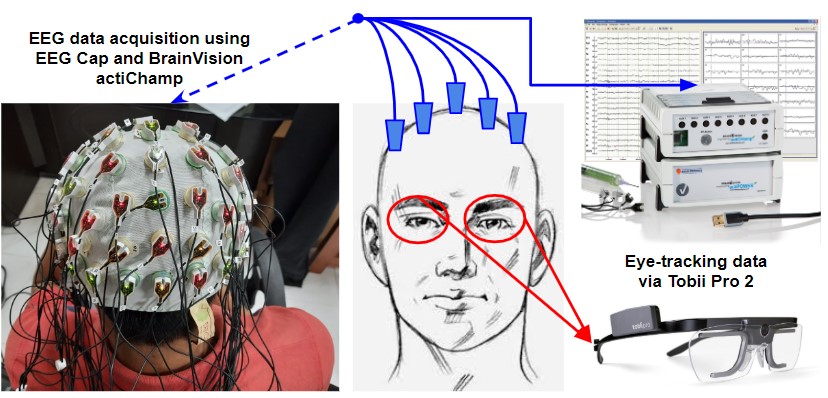}}\\
    %\subfigure[]{\includegraphics[width=0.48\textwidth]{f16}}
    \caption{Participants wearing EEG cap before the experiment. ActiCHamp\cite{actichamp} has been used in this work to record the EEG.Tobii eye-tracking glasses are used for capturing eye-movements} %(b) 2D and 3D map of the EEG electrodes as generated from EEGLab \cite{eeglab} mapped according to 10-20 system.}
    \label{f2}
\end{figure}
%Detection of informative shots/clips in a video sequence is a difficult task and a challenging problem.
In this section, we describe the experimental protocol and the proposed framework used for our work.
\subsection{Dataset Preparation}
\begin{figure*}[t]
    \centering
    \includegraphics[width=0.8\textwidth]{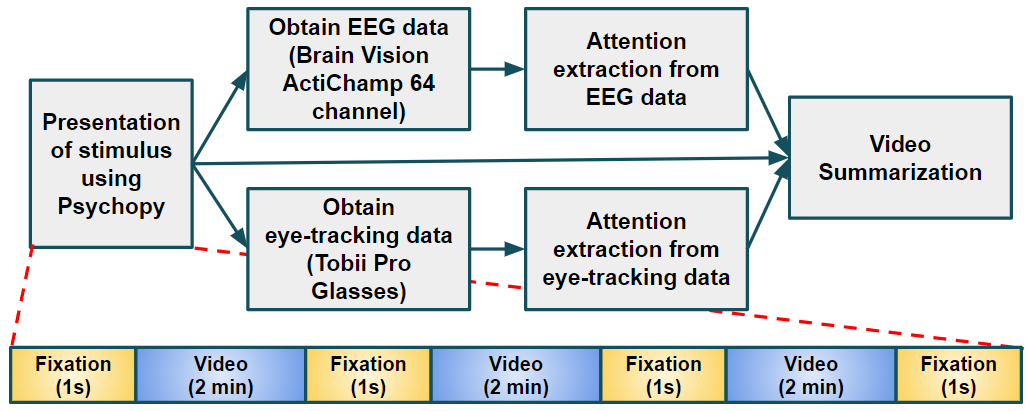}
    \caption{Experimental design for stimulus to be presented to the participants.}
    \label{f3}
\end{figure*}
\begin{figure*}[t]
    \centering
    \includegraphics[width=\textwidth]{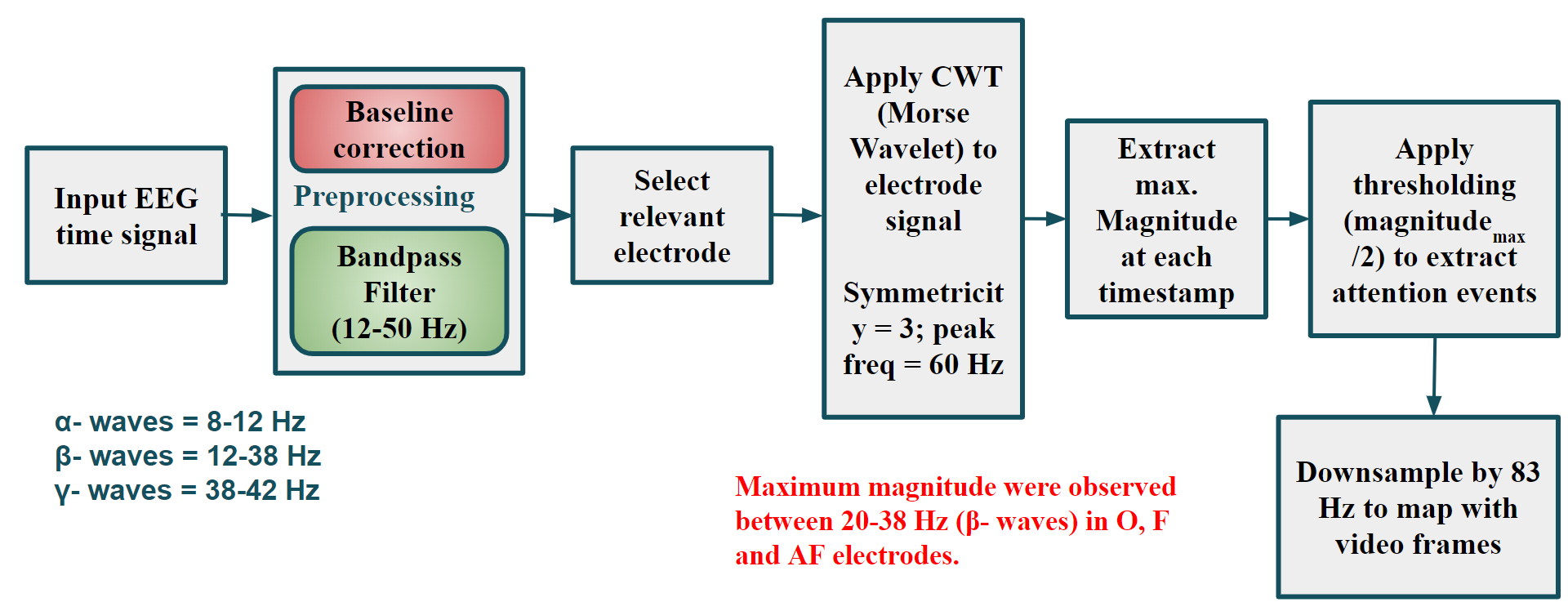}
    \caption{Attention extraction pipeline for the acquired EEG data. The bandpass filter is determined to extract EEG signals corresponding to frequency ranges covering $\alpha$, $\beta$ and $\gamma$ brain waves, the range of which is given in the figure \cite{miller2007theory}. The frequency of sampling for the EEG signal is 2500 Hz.}
    \label{f5}
\end{figure*}
To create the database for surveillance video, we recorded a 60-minute video of the student activities in a room at the Indian Institute of Technology, Delhi, with due permission from the students and the concerned authorities.

%\subsubsection{Participants}
%Students from Indian Institute of Technology Delhi participated in the experiment. All participants had a normal or corrected-to-normal vision with no known neurological impairments.

\subsection{Participant Screening}
Total 15 participants, with normal or corrected to normal vision actively participated in the experiment. All the participants were healthy with an average age of 25 years ($\pm$ 2.5 years). Each participant was assigned a different identification number for maintaining data confidentiality.

\subsection{Eye-tracking and EEG Equipment}
For our work, we want to acquire both the brain signals (EEG) and eye-tracking data from the participants when presented with stimuli. We have used BrainVision actiChamp with 64 channels for EEG signal acquisition. The electrodes were placed according to the international 10–20 system \cite{homan1987cerebral}. Brain Vision actiChamp \cite{actichamp} amplified the collected signals and EEG data was collected at a sampling frequency of 2500 Hz. The eye-tracking data was collected using Tobii Pro-Glasses-2 having sampling frequency of 100 Hz.
%The software that was used to analyse further the eye-tracking data are Tobii Pro Controller and Tobii Pro Analyser. 
\subsection{Stimulus Design and presentation sequence}
From the acquired video, short snippets of 2-minute videos were created, and an experiment was designed. Fig. \ref{f3} depicts the experimental protocol used for data collection. As observed from the Fig. \ref{f3} , we have presented the video snippet (2 min) from our database interleaved by a fixation screen. The fixation screen was essentially a grey screen with a fixation cross at the center displayed for 1s. The stimulus i.e.2 min video snippet was presented on a monitor via Psychopy software tool \cite{peirce2019psychopy2}, which is an application to create experiments in behavioral science. 

%For this tool, the users can write scripts in Python or construct experiments graphically using the Builder interface as provided by the software. 
%\subsection{Methodology}

%\subsubsection{Creating dataset}

\subsection{Experimental Protocol}
Before the start of the experiment, we described the experimental procedure to each participant. After getting the consent from the participant, he/she was asked to sit comfortably in front of a monitor screen which was fixed at a distance of 60-80 cm. The EEG cap was put on the head of the participant. We placed the electrodes on the cap. The contact between the scalp and electrode was made through the insertion of gel. The impedance limit in the actiChamp i.e. EEG amplifier, was kept at 10 k$\Omega$ to minimize the noise and better quality EEG signals. Fig. \ref{f2} shows participants prepared before the experiment wearing EEG caps. Once, all electrodes showed the impedance level below 10k$\Omega$; we made the participant wear the eye-tracking glasses. The calibration process was performed before the start of the experiment for all the participants to ensure useful eye-tracking data. We have used the 9-point calibration to ensure good quality data.
\begin{figure*}[t]
    \centering
    {\includegraphics[width=1\textwidth]{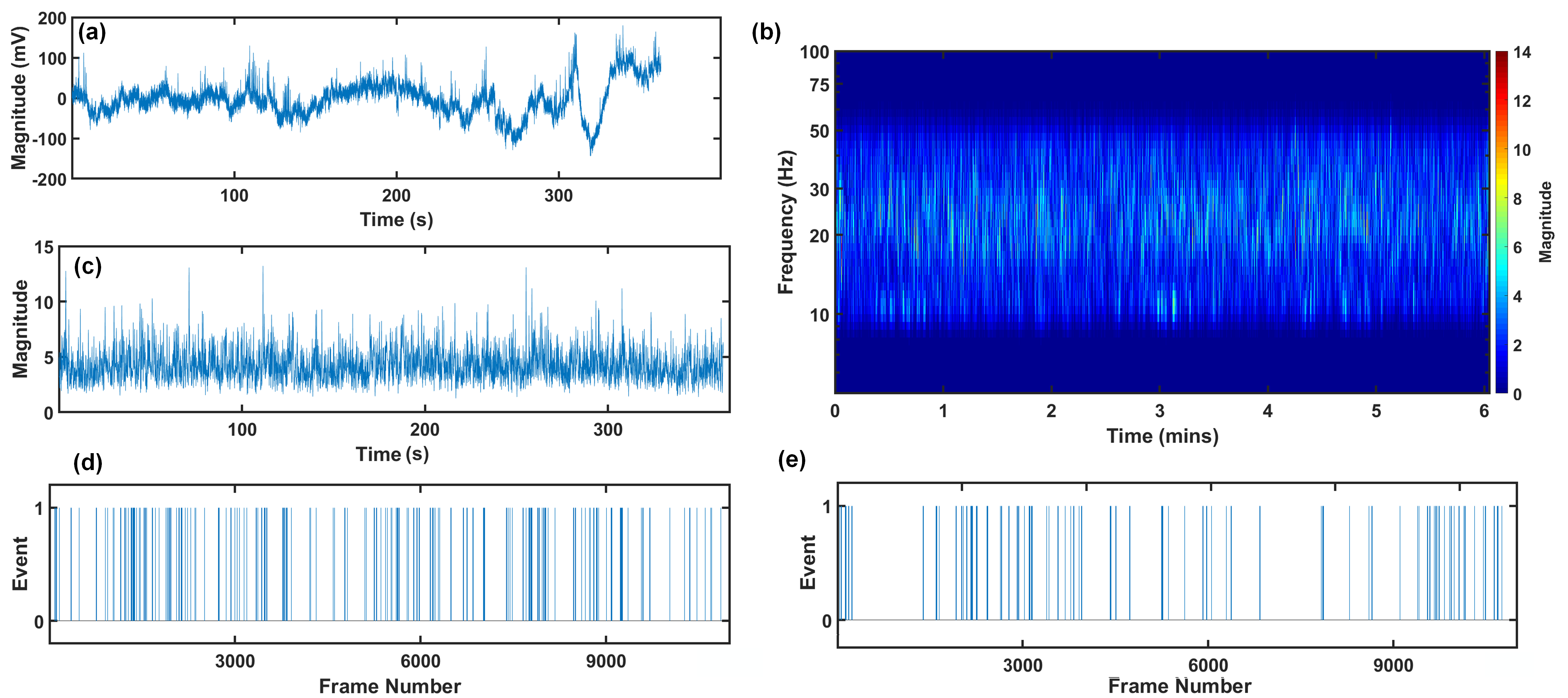}}
    %\subfigure[]{\includegraphics[width=0.5\textwidth]{image/f6_a.png}}\subfigure[]{\includegraphics[width=0.5\textwidth]{image/f6_b.png}}\\
    %\subfigure[]{\includegraphics[width=0.5\textwidth]{image/f6_c.png}}\subfigure[]{\includegraphics[width=0.5\textwidth]{image/f6_d.png}}\\
    %\subfigure[]{\includegraphics[width=0.5\textwidth]{image/f6_e.png}}\subfigure[]{\includegraphics[width=0.5\textwidth]{image/f6_f.png}}
    \caption{(a) EEG signal data from AF7 electrode before any pre-processing is done. EEG data is acquired at sampling rate of 2500 Hz (b) CWT applied to the band passed signal after baseline correction. (c) Maximum magnitude corresponding to each time stamp in the EEG signal. (d) Extracted attention events from the EEG signal (e) Extracted attention events from the EEG signal O1. The attention events obtained here give an estimate where the anterior frontal lobe identifies important events. Other electrodes showing some significant activity includes OZ, O1 and O2 (from occipital lobe); F1$-$F7 (frontal lobe); Fp1 and Fp2 (Pre-frontal lobe); and, AF3, AF4 and AF8 (from anterior frontal lobe). }
    \label{f6}
\end{figure*}
We piloted the experiment on a couple of participants by showing them the whole 60-minute recording. However, it was creating fatigue and was affecting the recording. Therefore, we have decided to cut the whole video in short snippets of 2 min, and the snippets were selected from the entire 60 min video judiciously. The video stimulus that composed of several snippets formed as a complete video of 6 min without losing any consistency was finally presented to each participant separately following the experimental protocol shown in Fig. \ref{f3}. The stimulus presentation time was about 10-15 min, and the experimentation, including EEG and eye-tracking setup time, took around 45-60 min. The video snippets consisted of both activity and non-activity frames, which are generally observed in the surveillance videos. The stimulus presentation was controlled using custom Python Script created for this work. The stimulus, which is essentially 2-minute short snippets from surveillance video of a 60-minute video database, is presented on the monitor via Psychopy. We have shown at least 3 2-minute video snippets to a participant. The participant was asked to carefully analyze the activities happening in the video, such as detecting suspicious persons or activities.

\section{Proposed Video Summarization Framework}
In this section, we will discuss the extraction of important frames from EEG and eye-tracking data, followed by a summarization based on these frames.

\subsection{Key-frame extraction pipeline from EEG signals}
\begin{figure*}[t]
    \centering
    \subfigure[]{\includegraphics[width=0.49\textwidth]{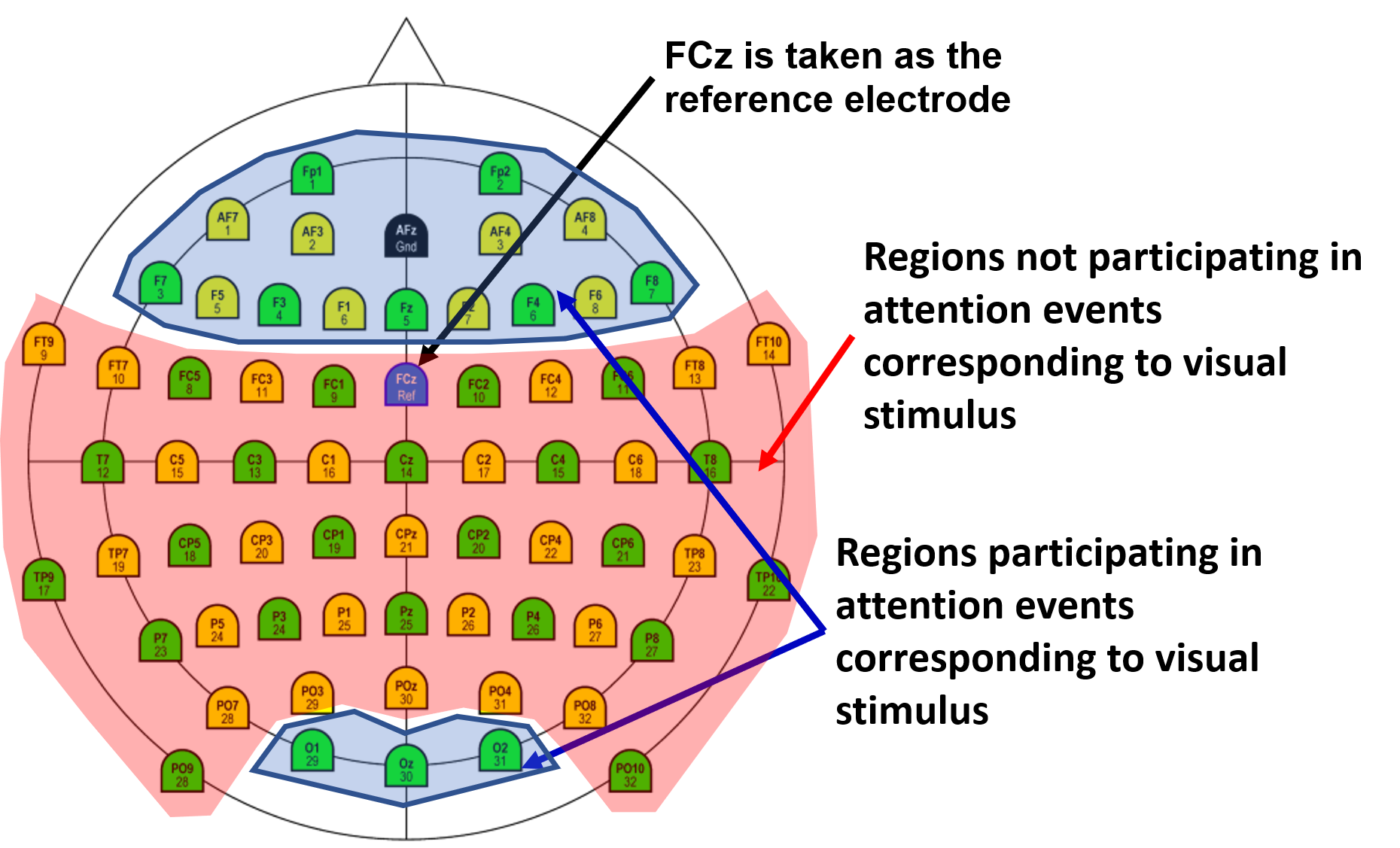}}
    \subfigure[]{\includegraphics[width=0.5\textwidth]{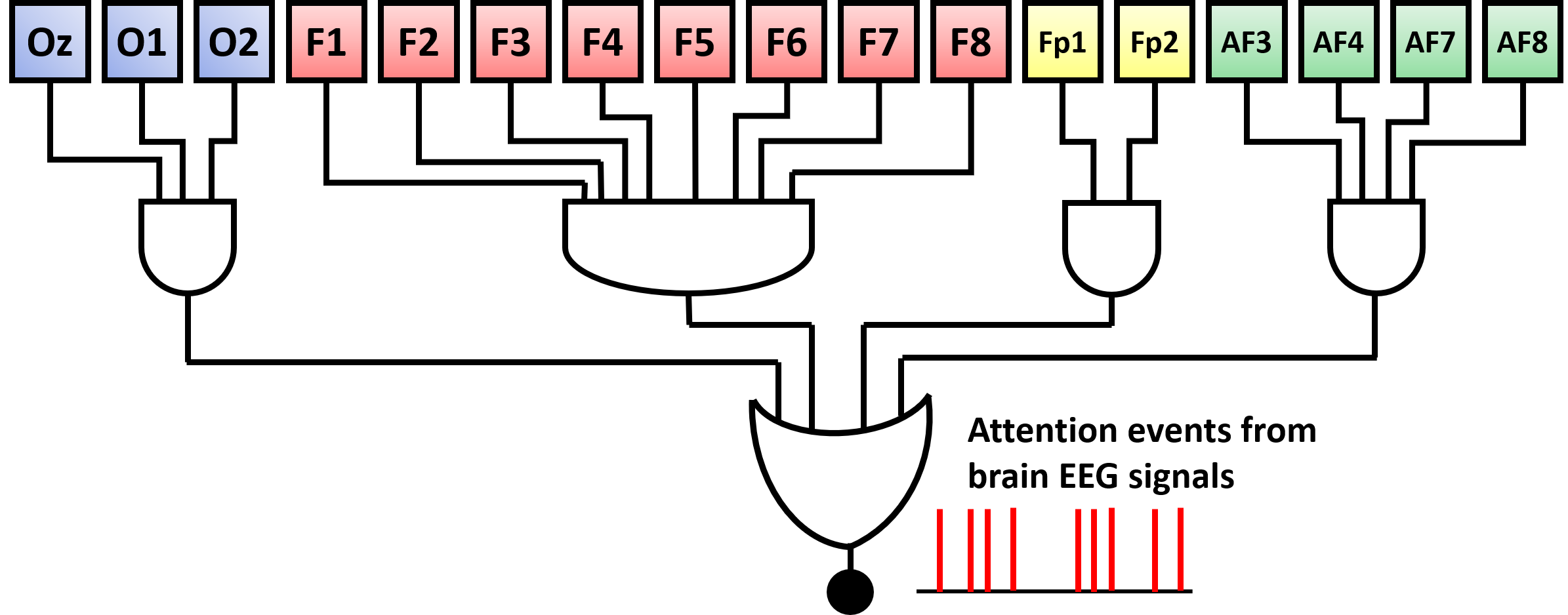}}
    \caption{(a) EEG electrodes which exhibit activities corresponding to the visual stimulus. The region shaded in blue shows the area of the brain from which attention events are extracted. The region in red did not show significant activity. (b) Overall event extraction pipeline for EEG signal analysis. Electrodes color coded in same color are clustered together as they approximately belong to one region of the brain.}
    \label{f7}
\end{figure*}
Among 15 participants (9 males, 6 females), 2 participants moved a lot while observing the stimulus. This has affected the data and hence we have only considered the data of remaining 13 participants. We are more interested in knowing where do the participants focus or attend in the video (important frames according to them). Therefore, we have extracted the attention-based brain signals through a pre-processing pipeline from the collected EEG data, as shown in Fig. \ref{f5}. All the steps mentioned in the attention extraction pipeline are executed in MATLAB. The EEG data (signal from AF7 is shown in Fig. \ref{f6}(a)) acquired is first pre-processed using a baseline correction to remove noise from the data and then passing the signal through a bandpass filter with a bandwidth of 12-50 Hz. The baseline correction is done via EEGLab \cite{eeglab}. EEGLab is an interactive MATLAB toolbox that combines independent component analysis (ICA), time/frequency analysis, artifact rejection, and event-related statistics to analyze continuous and event-related EEG, MEG, and other electrophysiological data \cite{eeglab}. The bandwidth is selected, such that the resulting filtered signal captures $\alpha$, $\beta$, and $\gamma$ waves. A continuous-wavelet-transform (CWT) is then applied to the signal from each electrode. Continuous wavelet transform provides a scheme for the interpretation of a signal, $x(t)$ while working with a single, possibly complex-valued wavelet function, $\psi(t)$. The time-scale  decomposition of a signal is given by:
\begin{equation}
    S(a,b) = \frac{1}{|a|}\left\vert\int_{\infty}^{\infty}x(t)\psi^*\left(\frac{t-b}{a}\right)dt\right\vert^2
\end{equation}
which expresses the energy of the signal at any scale and time. Here , $\psi^*(\cdot)$ denotes the complex conjugate of $\psi(\cdot)$. Fig. \ref{f6}(b) shows CWT performed on band pass filtered signal of AF7 electrode, over the entire frequency range. For this work, we only need CWT of the signal after band pass filtering(BPF). Using CWT, we break down the signal into smaller time stamps and perform frequency transform (FT) to each timestamp. Thus, each time stamp will give the frequency and the magnitude corresponding to each frequency in that timestamp (Fig. \ref{f6}(c)). After obtaining the maximum magnitude, a threshold value is selected. For our work, we have kept the threshold value to be half of the maximum magnitude in that timestamp. Now, those time stamps having magnitude more that the threshold value are considered to be important and signifying the frames where human attention is observed. Thus, we are naming these time stamps as attention events. Now, our video is recorded at 83 fps. Therefore, the attention events are then down-sampled to 83 to map the events corresponding to the video frames. Fig. \ref{f6}(d) shows the extracted attention events corresponding the signal from AF electrode and the corresponding frames. Similarly, attention events extracted for O1 electrode as shown in Fig. \ref{f6}(e). This process is followed for the electrodes from occipital and frontal regions of the brain.

After processing the EEG signals with the attention extraction pipeline, we could observe that there is some significant activities in the frontal and occipital lobe, when the participants were presented to the stimulus. Frontal lobe of the brain is often associated with decision making, reasoning and complex calculation \cite{gil2008eeg,kyathanahally2016realistic,alarcao2017emotions}. The electrodes, which showed some activity corresponding to the visual stimulus, are highlighted in Fig. \ref{f7} (a). We consider the electrodes showing activity in these regions and cluster them based on the region to which the electrodes belong to. The illustration of the clustering of electrodes from the region of interest can be observed in \ref{f7} (b). In this illustration, we show how the overall extraction pipeline is able to come up with an cumulative attention event corresponding to the visual stimulus. We propose to use the logical AND operation on the electrodes coming from the same regions to reduce noise. However, the electrodes which are away from each other and lying in different regions also contribute to attention. Therefore, to have the combined effect of it, logical OR operation is performed between the electrode clusters from different regions of the brain (which show some activity during stimulus) to give us a overall attention event from the brain. This logical combination fetches us the final attention events. The pipeline is shown in Fig. \ref{f7} (b). The procedure of event extraction is repeated for all the participants.

\begin{figure}[t]
    \centering
    \includegraphics[width=0.49\textwidth]{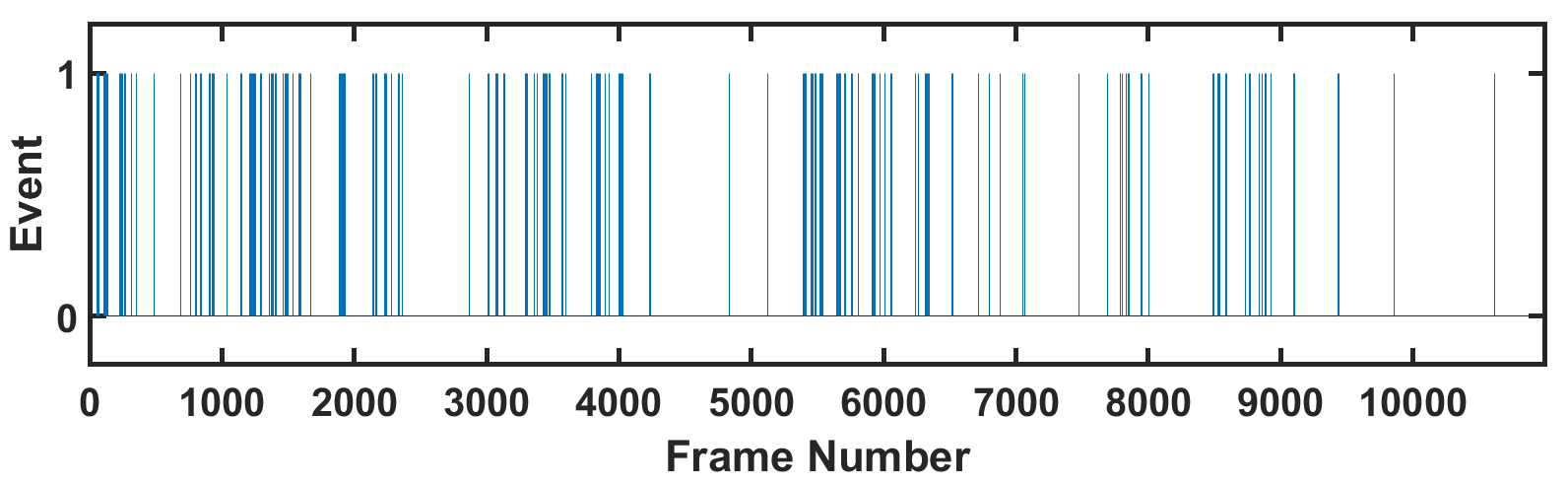}
    \caption{The key-frames extracted from the EEG signal of a single participant obtained from the overall event extraction pipeline (as shown in Fig. \ref{f7} (b)). }
    \label{f8}
\end{figure}

Fig. \ref{f8} shows one example of extraction of attention events from  the EEG signal of a single participant. Each spike in the plot shows the activation events in the brain when the video was being played for the participants. The spikes therefore corresponds to important events from the video as judged by the brain.

\subsection{Key-frame extraction pipeline from eye-tracking data}
\begin{figure}[t]
     \centering
    \includegraphics[width=0.5\textwidth]{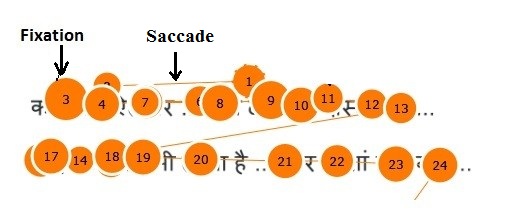}
     \caption{The circles are Fixations and the distance between two consecutive fixations are sccades. Each fixation is numbered based on the order it was captured. The radius of the circle increases as a function of fixation duration. \cite{ralekar2017unlocking}.}
     \label{chetan}
 \end{figure}

The eye-tracker provides the raw eye gaze information. From the eye-gazes, we need to extract eye-tracking features. There are two basic eye-tracking features viz. fixation and saccade (Fig. \ref{chetan}). Fast and simultaneous movement of eyes is called a saccade. It is the rapid eye movement between fixations to move the eye-gaze from one point to another. On the other hand, fixations are the points where the eyes are relatively stationary. A fixation comprises of slower and minute movements that stimulate the eye to align with the target and avoid perceptual fading. Fixations provide us an idea about the location's information processing where the eyes are relatively stable \cite{rayner1998eye}. Thus, while extracting key-frames, fixations are relatively crucial as compared to the saccade. We have used the velocity threshold identification (I-VT) algorithm to obtain the eye-tracking features from the raw eye-tracking data\cite{salvucci2000identifying}. 

%\subsubsection{Extracting overall attention events from EEG signal}
%After obtaining the attention events from all the relevant electrodes, we apply the event extraction pipeline to all the resultant signals. %In this pipeline, as EEG signals have a poor spatial resolution. Therefore, we are not able to locate the areas responsible for a particular activity that precisely. However, 
%When a participant was presented with the stimulus video, we could observe some activity frontal and occipital lobes. 

\begin{figure}[t]
    \centering
    \includegraphics[width=0.5\textwidth]{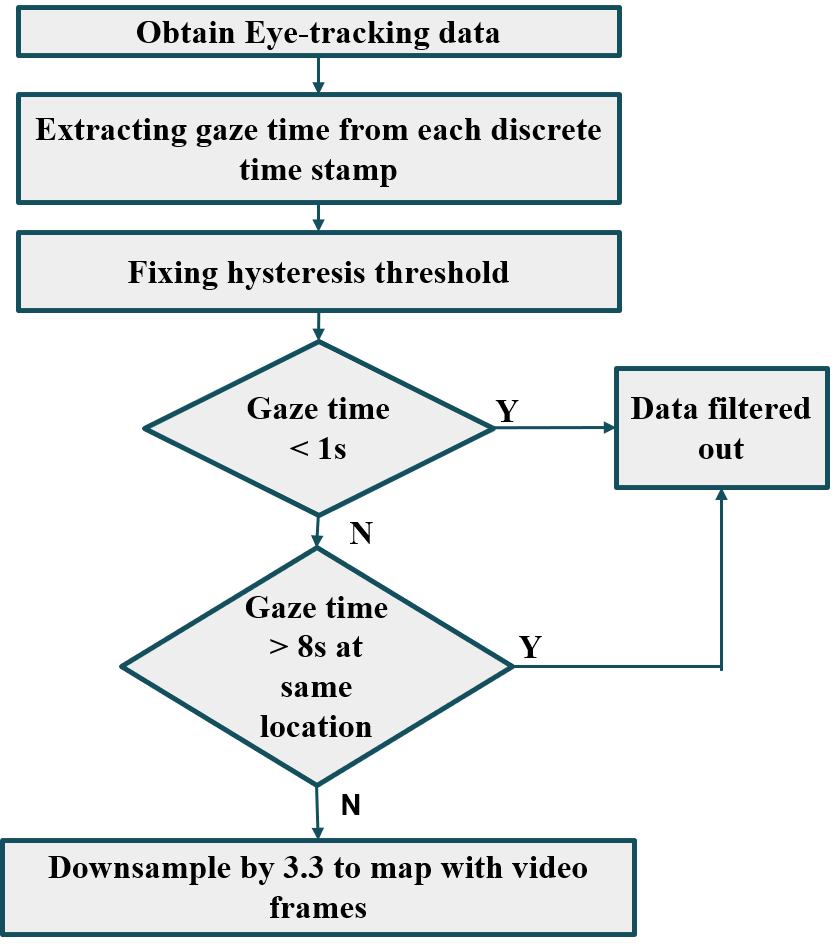}
    \caption{Attention event extracting pipeline for data obtained from eye-tracking. The sampling time for the Tobii Pro 2 is 100 Hz.}
    \label{f9}
\end{figure}
\begin{figure}[t]
    \centering
    \includegraphics[width=0.5\textwidth]{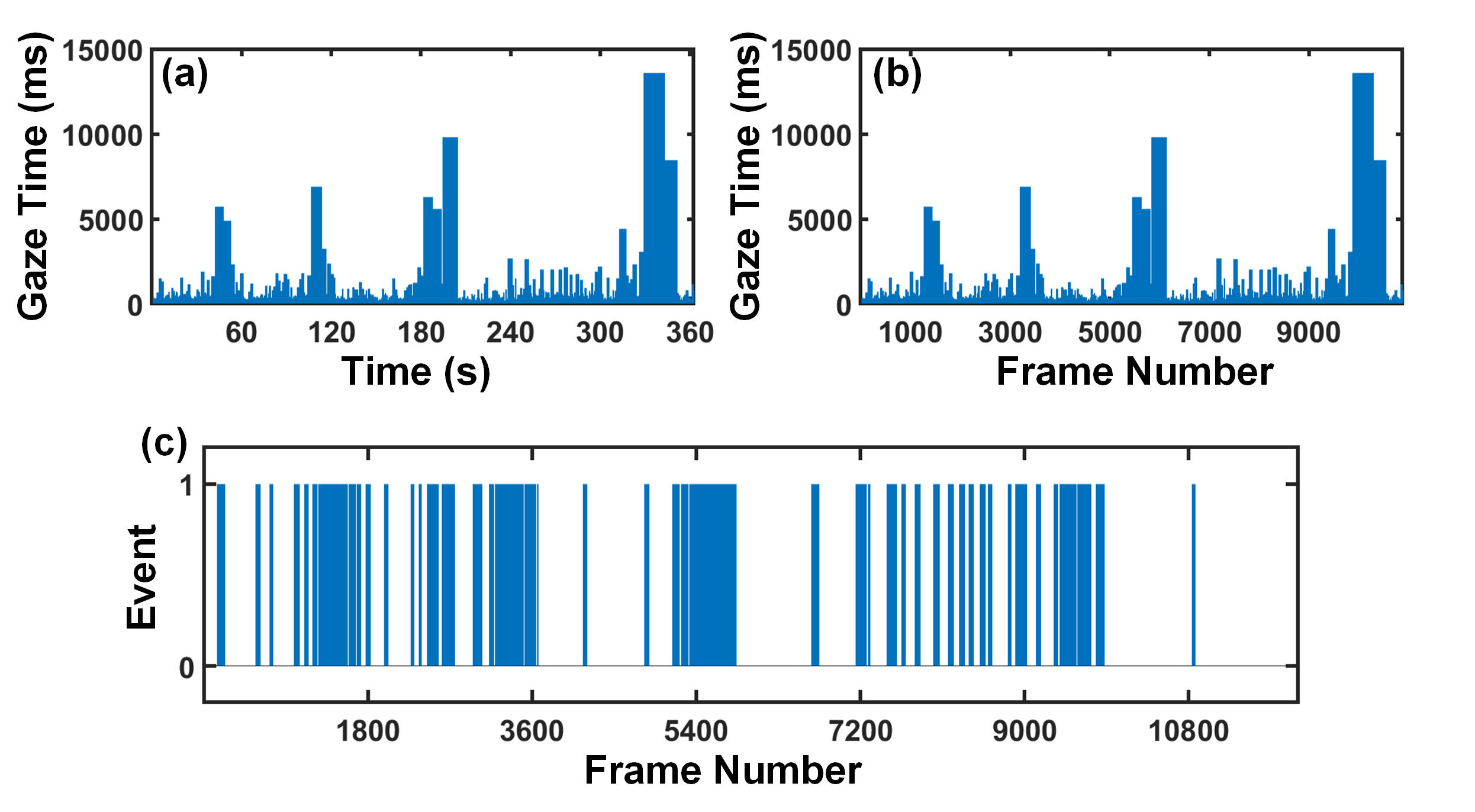}
    \caption{(a) Eye-tracking data obtained from Tobii Pro 2. The sampling frequency is 100 Hz which is down-sampled by 3.3 Hz to map with the video frame. (b) Down-sampled data which can be mapped with the video frame. (c) Resulting attention events extracted by removing saccade and boredom data from the down-sampled data.}
    \label{f10}
\end{figure}
\begin{figure}[t]
    \centering
    \includegraphics[width=0.5\textwidth]{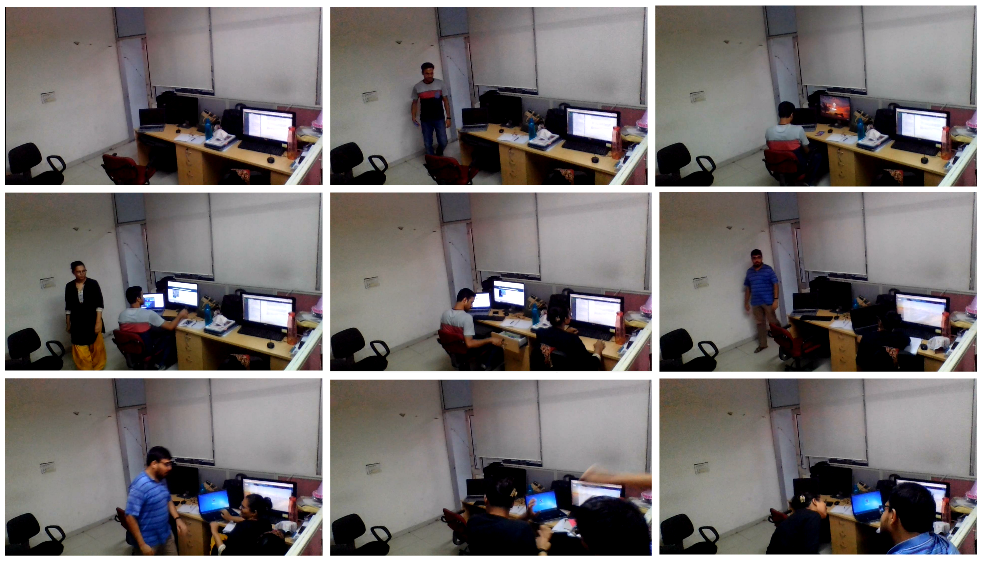}
    \caption{Extracted key-frames from the 6 minute long video showing how from EEG and eye-tracking attention data we are able to summarize the video. The first frame is captured since the participant gets familiarized with the video environment in the first few seconds and thus concentrates more.}
    \label{f13}
\end{figure}

The data is acquired from the eye-tracker at the sampling rate of 100 Hz (Fig. \ref{f10} (a)). Therefore, each timestamp (1s) corresponds to 100 points in the raw data obtained from the instrument. The data is then down-sampled by 3.3 to map to the video frames ((Fig. \ref{f10} (b))). A hysteresis threshold is fixed where gaze time ($t_{gaze}$) less than 1000 ms (1 s) is considered as a saccade, and $t_{gaze}$ more than 8000 ms (8 s) is considered as boredom, and the corresponding data is discarded. We have included the condition for boredom since surveillance videos tend to be sporadic in terms of events, and the subject might get bored while observing the video. Gaze time between 1s to 8s is the fixation data from which the attention events are extracted. Fig. \ref{f10} (c) shows the attention extracted output from the eye-tracking data.
If we look at the attention events extracted from EEG and eye-tracking data, we can see that the events obtained are more or less the same. It is essential to understand that eye-tracking gives us more key-frames, and there might be redundant frames. The more informative frame might also trigger brain activity. Hence, if we select the frames which are commonly extracted from both EEG and eye-tracking attention pipeline, we can discard unwanted or redundant frames. Therefore, to obtain the overall attention events corresponding to both the brain and the eyes, we decided to perform a logical AND operation on the event extracted data. The resulting key-frames are shown in Fig. \ref{f13} are used for video summarization.
%\begin{figure}[t]
    %\centering
    %\includegraphics[width=0.35\textwidth]{f11}
    %\caption{EEG and eye-tracking (ET) attention data as obtained after pre-processing}
    %\label{f11}
%\end{figure}

%\section{Neural Network Framework}
%\begin{figure*}[t]
    %\centering
    %\includegraphics[width=\textwidth]{f12}
    %\caption{Proposed bio-inspired video summarization neural network framework (BioNN). V = \emph{valid} which indicates that the keyframe is important whereas I = \emph{invalid} which indicates keyframe has no important data that can be summarized.}
    %\label{f12}
%\end{figure*}

%\subsubsection{Key-frame Extraction}
%The frames of surveillance video are extracted at the attention timestamps using above procedure. To reduce the computation time for video
%segmentation, fraction of the frames can be used as input to frameworks similar to that in VSUMM. Being supervised techniques, the human signal based information can make the approach more bio inspired driven and the computation cost would be decreased by a significant margin as already data is summarized once. 

\subsection{Video Summarization}
We are collecting  both EEG and eye-tracking data which is acquired at a sampling frequency of 2500Hz and 100Hz respectively. Therefore, it is important to synchronized the data from both technologies to video frame rate. Thus,  
in this step, attention events at each timestamp are extracted from EEG and eye-tracking data processing over the entire presented video. The EEG and eye-tracking data are synchronized to video frame rate. The frames, where we observe high attention (i.e through EEG and eye-tracking data processing), are marked as key-frames (frame of interest). The collection of key-frames would furnish a refined raw video form without losing much information.
{In general, the ground truth for any task is generated by intervention of human experts. In our case, we have taken the help of expert to identify valid and invalid key-frame for summarization task. The total video is broken down into different frames and shown to the expert to classify those frames into valid and invalid.}

\begin{figure}
    \centering
    \includegraphics[width=0.30\textwidth]{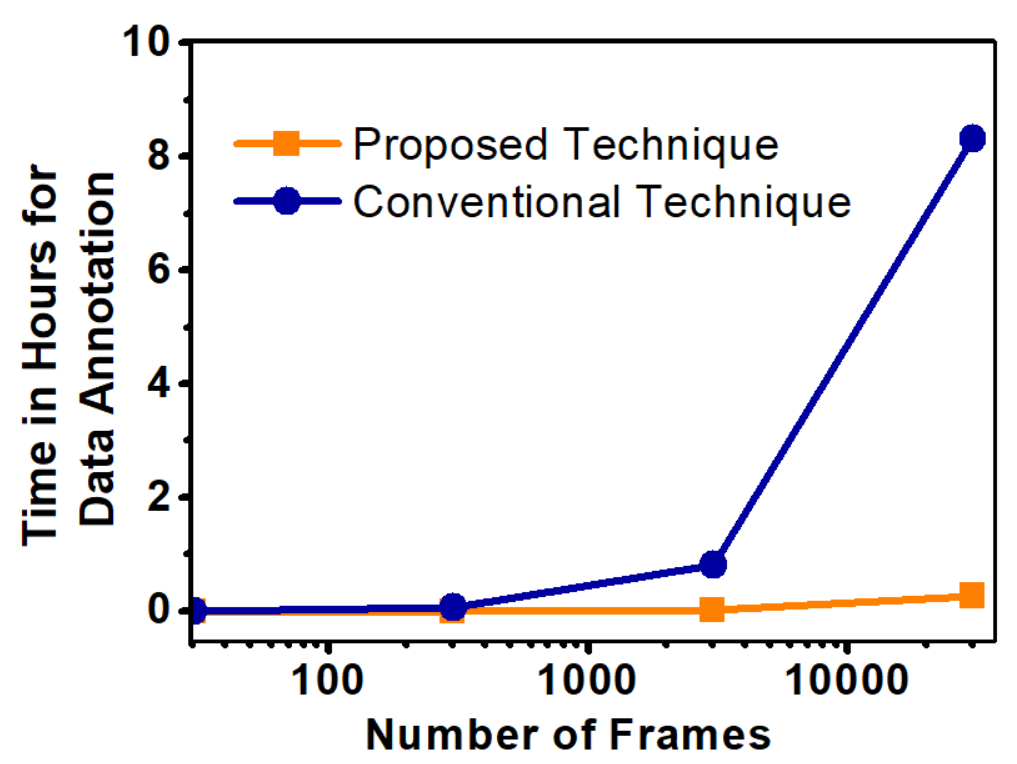}
    \caption{Qualitative representation of merit of proposed method over conventional method, where horizontal axis represents number of frames and vertical axis time required to annotate in hours.}
    \label{fig:prop_conv_gtl}
\end{figure}
\section{Results and Discussion}

\subsection{Advantage of using both EEG and Eye-tracking}
It might be argued that why do we need to consider both the EEG and eye-tracking data. In this context, we performed an analysis where we extracted the key frames using only EEG signals, only eye-tracking signals and the proposed scheme where both EEG and eye-tracking data has been considered fr video summarization.  Precision and recall are two important model evaluation metrics when performing information retrieval and classification in pattern recognition processes. In principle, a high precision essentially means that the framework can extract relevant information from the given dataset. On the other hand, a high recall factor correlates the matched key-frames with the actual ground truth. In other words, how efficient it is to obtain the ground-truth (actual required key-frame). Therefore, the comparison is quantified using the following two equations:
\begin{equation}
    Precision=\frac{N_{matched}}{N_{extracted}} 
\end{equation}
\begin{equation}
    Recall=\frac{N_{matched}}{N_{ground-truth}}
\end{equation}

where, N$_{matched}$ = number of frames extracted are matching with ground-truth, N$_{extracted}$ = number of video frames extracted by the framework under study and N$_{groudtruth}$ = number of frames corresponding to the ground truth. As our dataset and ground truth are made by us, { we have tried to compare precision where algorithm had given highest recall.}
The performance metrics is given in Table \ref{t2}. From the table we can observe that while the methodology using only EEG gives a better compression factor as compared to methodology using only eye-tracking, the ratio between the number of key-frames extracted and the number of matched key-frames is higher as compared to the proposed algorithm. Moreover, if we look at the precision, recall and F-value parameters for the methodologies, we can observe that using only EEG for video summarization has 31\% less precision and 1\% less recall as compared to the proposed methodology. Although using only eye-tracking has a very high recall value, the precision is 83\% less than methodology which uses both EEG and eye-tracking data for video summarization. This analysis shows that using both EEG and eye-tracking for video summarization not only yields higher precision and recall values, but also higher compression factors and better accuracy.

\subsection{Reduced manual efforts in ground truth generation}
{ The conventional way of generating the ground truth is to break the total video into different frames and show those frames to the expert. The expert will use the domain knowledge and the label each frame as valid or invalid. This method of ground truth generation is applied for any task. If we assume a human expert annotate video at 1 FPS (or lower), he will take around 2 hours to annotate 1200 frames and 7-8 hours for 10000 frames. A huge amount of time is invested in labeling the video frames. Most of the videos are recorded at higher sampling rate (i.e. temporal resolution) which in turn increases the amount of frames to be labeled. To avoid the manual efforts, most of the Video summarization frameworks work on low temporal resolution i.e. the original video is downsampled by using a downsampler as a pre-processing block. When we use downsampling, the number of frames which are fed to the summarization framework will be less as compared to the original video. However,such downsampling of the data may result in the loss of key-frames and it be interpreted as loss of data \cite{de2011vsumm}. Indeed, the limitation in such a case is high temporal resolution data (all video frames), which is computationally very expensive for a deep learning architecture to work on. In such scenario, the proposed approach has a distinct edge over other summarization models. In proposed framework, we show complete video to the participant and we extract the key-frames from the whole video, instead of downsampled video. One can annotate data at very high resolution with the capabilities of EEG and Eye tracking easily. There is no need of breaking the video into several frames and subsequent investigation of each frame to classify it as valid and invalid. Using our approach, expert will only see the complete video and annotation will be done using EEG and Eye-tracking signals. This approach will significantly reduce the manual efforts in annotating every video frame.  The Fig. \ref{fig:prop_conv_gtl} qualitatively shows the merit of video annotation with our proposed method and conventional Method.}

%\st{Here in our setup due to our limitation on subjects being college students, the presented video is chosen as surveillance video being a simpler task. Hence all human subjects had shown similar attention. But, if this methodology can be extended to specific tasks like watching a movie, different persons can give different summary according to their emotional tendencies. Using conventional process, it would be very difficult to annotate a video according to a person, which can be done much easily with proposed methodology. In an other scenario, when specific video is shown to professionally specialized persons, it will show a path to understand human cognition capabilities to make a efficient system, Hence, the proposed method moves us towards explainable AI, where we can understand the human attention to more bio-realistic networks.}}

\subsection{Comparative Analysis}
\begin{figure}[t]
    \centering
    \includegraphics[width=0.48\textwidth]{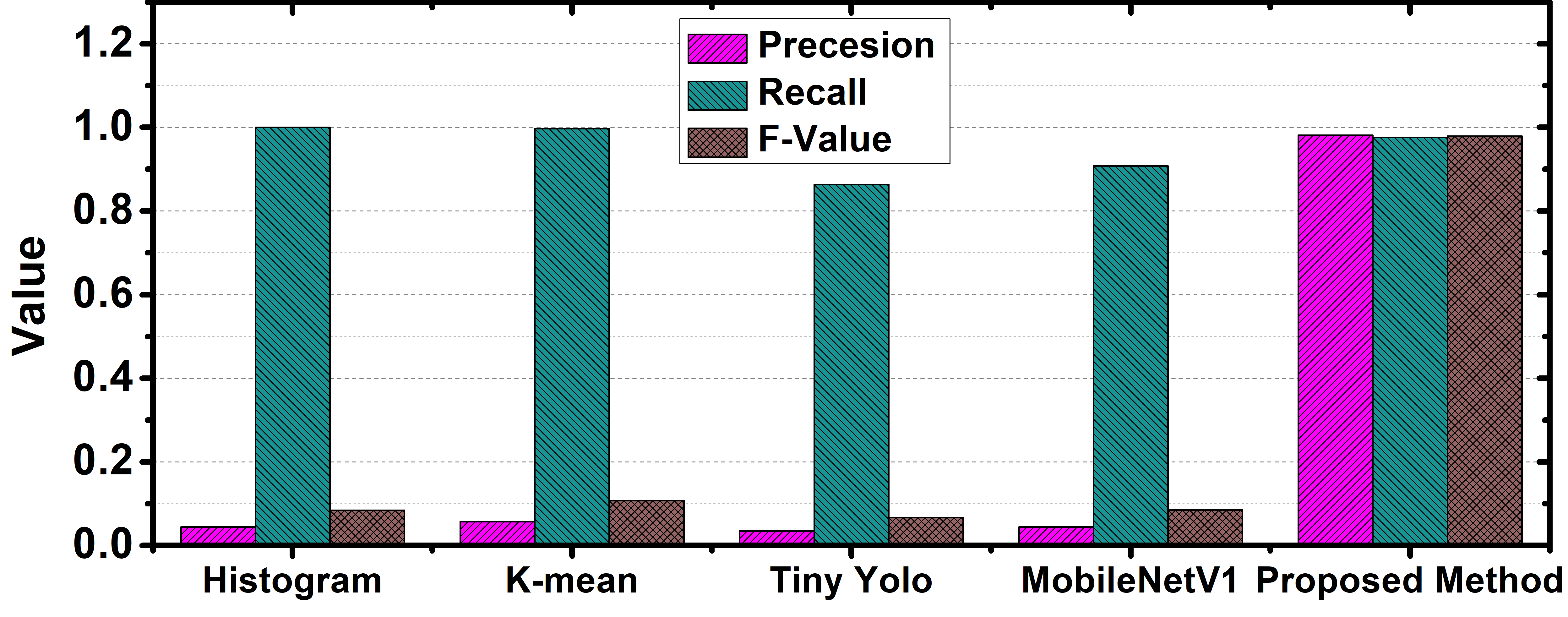}
    \caption{Keyframe extraction benchmarks highlighting the performance of different keyframe extraction frameworks.}
    \label{f18}
\end{figure}

%\section{Comparison and Benchmarking}
We have compared our proposed scheme with two low-level feature frameworks, Histogram \cite{almeida2012vison} and Kmeans \cite{kmeans}, and two object detection based AI (artificial intelligence) frameworks, Yolo3 and MoileNetV1 \cite{mob}.  Proposed framework results shown are data received by the best of 13 participants data. For low-level feature frameworks, we have selected parameters with the highest recall. Table \ref{t1} shows the performance benchmarks for different frameworks considered in our paper. The compression factor mentioned in the table indicates how much the total video dataset is compressed. From the table, we can observe that the proposed framework has the highest precision, which denotes that the algorithm returned more relevant results than irrelevant ones. We also observe that the proposed framework's recall factor is $\sim$ 0.97, which indicates that the algorithm returned most of the relevant results. It is to be noted here that the proposed results are limited to our dataset. Since there is a balance obtained between the precision and the recall, we measure the performance of the proposed method using the F-scale values. From Fig. \ref{f18} it can be clearly observed that F-value is highest for the proposed methodology thereby signifying that it is able to extract the key-frames with better accuracy.
%Furthermore, the proposed framework can compress the data by a factor of 0.96 while keeping a high precision ($\sim$ 0.98) and recall rates ($ \sim $ 0.97). This result demonstrates that using human perception; we are able to extract the key-frames in an efficient and precise manner. 
Similarly, we also observe that the detection percentage (portion of correctly extracted key-frames) is also better for our proposed system as compared to other frameworks.

\begin{table}[t]
    \centering
    \caption{Ablation study data showing the advantage of using both EEG and Eye-tracking in video summarization}
    \label{t2}
    \begin{tabular}{ccc}
    \hline
    \textbf{Methodology} & \textbf{Compression Factor} & \textbf{Detection Percentage}\\
     %& \textbf{Factor} & \textbf{Percentage}\\
    \hline
    \textbf{EEG}  & 0.95 & 68.7\\
    \textbf{Eye-tracking (ET)} & 0.77 & 15.27\\
    \textbf{EEG + ET} & 0.96 & 98.15\\
    %Ground Truth & 10893 & 380 & 380 & - \\
    \hline
    \end{tabular}
\end{table}
\begin{figure}[t]
    \centering
    \includegraphics[width=0.48\textwidth]{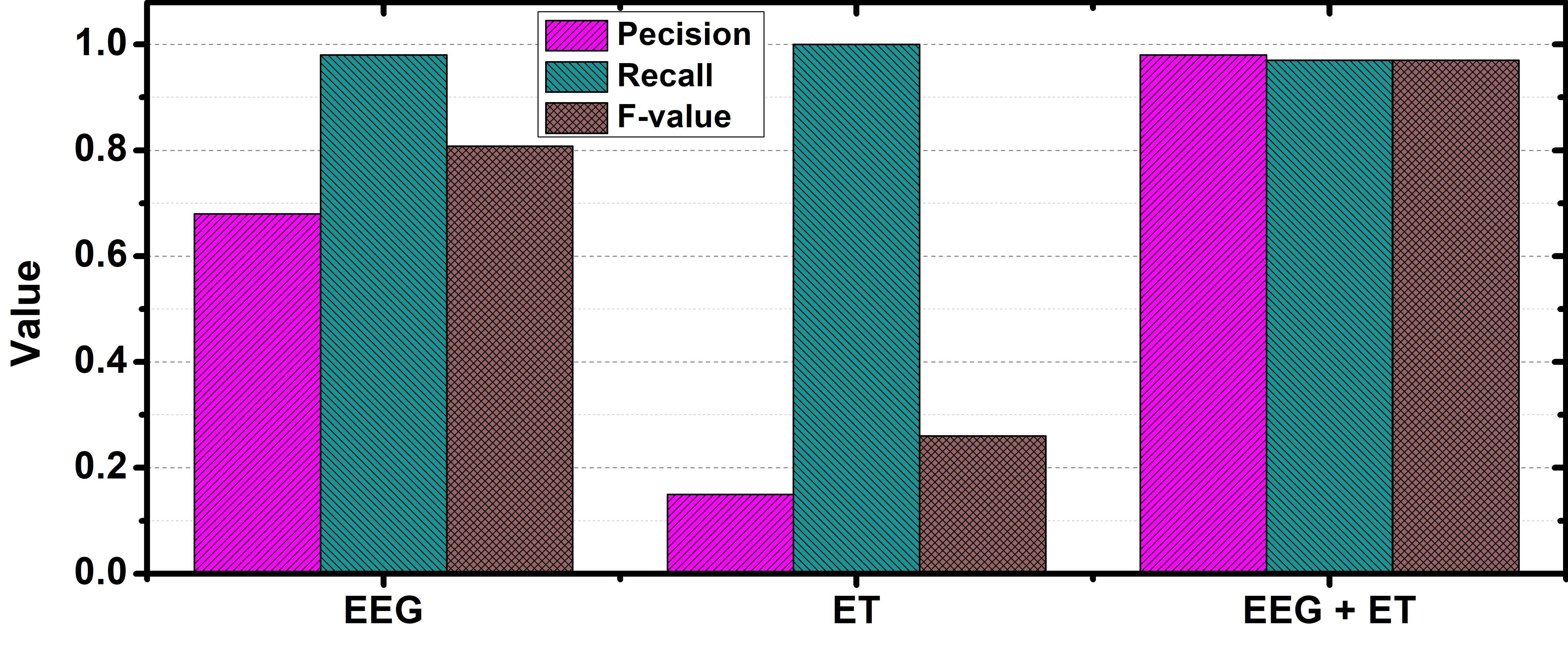}
    \caption{Keyframe extraction benchmarks highlighting performance when considering video summarization using only EEG, only eye-tracking (ET), ground truth (GT) and using the proposed scheme where both EEG and eye-tracking is used.}
    \label{f19}
\end{figure}
 
\begin{table}[t]
\caption{Compression Factor and Detection percentage (DF) comparison with other frame works.}
\label{t1}
\begin{tabular}{ccc}
\hline
\textbf{Methodology} & \textbf{Compression Factor} & \textbf{Detection Percentage} \\ \hline
\textbf{Histogram \cite{almeida2012vison}}       & 0.2  & 4.40     \\ 
\textbf{Kmeans \cite{kmeans}}          & 0.38   & 5.66  \\ 
\textbf{Tiny Yolo \cite{mob}}           & 0.13                       & 3.47   \\ 
\textbf{MobilenetV1 \cite{mob}}     & 0.28                        & 4.44         \\ 
\textbf{Proposed Method} & 0.96                       & 98.15      \\ \hline
\end{tabular}
\end{table}

\subsection{Discussion}
When it comes to achieve best possible performance in computer vision task, more often deep learning architectures form the first choice among researchers. However, such architectures needs more parameters resulting in increased computational complexity. Moreover, the internal functioning is very hard to understand. Humans are hardly able to understand how a system is achieving best possible performance. The proposed framework is not a deep learning-based approach; hence it requires very few parameters. Furthermore, the proposed summarization framework is easy to implement and reasoning can be easily understood as inputs from humans are involved in the design of the framework. Humans being the ultimate analyzer of the videos, input in the form of perceptually driven cues make the system explain its decision to humans and makes the system self explainable. The proposed work is one of its own kind which tries to get some cues through EEG and eye-tracking experimentation which can help in key-frame extraction. Perhaps, the proposed architecture can be looked as a proof of concept where the cues from perceptual and attentive behavior of humans can be utilized to enhance the system's performance. However, looking at the recent trends and volume of data, we plan to extend this idea to more deeper architectures wherein we can utilize these cues as an auxiliary input to enhance the system's performance. Thus, artificial systems can be made more intelligent when augmented with inputs from humans.

\section{Conclusion}
In this work, we have proposed a video summarization framework which is inspired by visual attention and perception. The use of EEG and eye-tracking technology give us some insights about how do humans perceive, analyze and interact with the environment. We have used these technology to extract attention events which has ultimately fetched key-video frames. Using these extracted key-frames, a video is summarized by $\sim$96.5\% while maintaining higher precision ($\sim$0.98) and high recall factors ($\sim$0.97). Moreover, we have analysed the effectiveness of the use of both EEG and eye-tracking for video summarization and concluded that using the proposed mechanism, we obtain high compression factor and accurate results.This work can be extended to augment the deep net based video summarization framework where the key-frames extracted using our approach can be used as supervisory input. This will not only enhance the performance but also make these architectures perceptually driven. In this way we could achieve, a human-machine collaborative video summarization framework to provide a holistic model to compress a large volume of video datasets, which is otherwise a very complicated process.
%\begin{table}[t]
%    \centering
%   \caption{Qualitative result of summarized videos from the dataset}
%    \label{t1}
%    \begin{tabular}{|c|c|c|c|}
%    \hline
%    & \textbf{Video} & \textbf{Summarized Video} & \textbf{Reduction} \\
%    & \textbf{Length (s)} & \textbf{Length (s)} & \textbf{(\%)} \\
%    \hline
%    Stimulus & 360 & 5 & 98.61 \\
%    Complete Dataset & 3600 & 121 & 96.63 \\
%    \hline
%    \end{tabular}
%\end{table}

%\appendix[Frequency Mapping of brain signals to obtain relevant electrodes]
%To obtain the relevant electrodes, we mapped an approximate ground truth of the 6 minute video with the attention events as obtained from all the EEG electrodes. If we look at the low power spectral density plot for the pass frequency of 6 Hz to 50 Hz (Fig. \ref{f17}), we can see that the electrodes or the area of the brain showing maximum activity are the ones that were obtained as relevant in the ground truth event mapping technique. This validates the use of O, F, Fp and AF electrode data used for training the neural network.
%\begin{figure}[t]
%    \centering
%    \subfigure[]{\includegraphics[width=0.5\textwidth]{f17_1}}\\
%    \subfigure[]{\includegraphics[width=0.5\textwidth]{f17_2}}
%    \caption{EEG spectral frequency map for the frequency range 6 Hz to 50 Hz. Red area shows maximum activity during 6 Hz and 20 Hz. (b) Low power density plot for the EEG data of 63 electrode with 70\% accuracy.}
%    \label{f17}
%\end{figure}
\section*{Acknowledgements}
%The authors would like to thank Dr. Tapan Gandhi for letting us use his lab facilities and providing us with valuable suggestions, Chetan Ralekar for helping us set up the experimental framework and helping with the EEG and eye-tracking data acquisition. 
The authors would like to thank all the participants who took part in the experiment.
\ifCLASSOPTIONcaptionsoff
  \newpage
\fi



%\bibliographystyle{IEEEtran}
\bibliography{ref.bib}
\bibliographystyle{IEEEtran}
%\begin{thebibliography}{1}


%\end{thebibliography}


\begin{IEEEbiography}[{\includegraphics[width=1in,height=1.25in,clip,keepaspectratio]{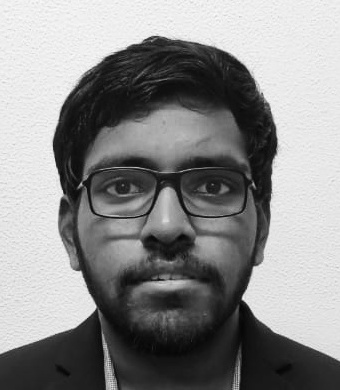}}]{Sai Sukruth Bezugam}
received B.Tech degree in Electronics and Communication from Keshav Memorial Institute of Technology, Hyderabad, India in 2018. He is currently enrolled as a MS (Research) student in the Department of Electrical Engineering, Indian Institute of Technology (IIT) Delhi, India. His research interests include neuromorphic engineering and computational neuroscience applications.
\end{IEEEbiography}

%\vspace{-5cm}
\begin{IEEEbiography}[{\includegraphics[width=1in,height=1.25in,clip, keepaspectratio]{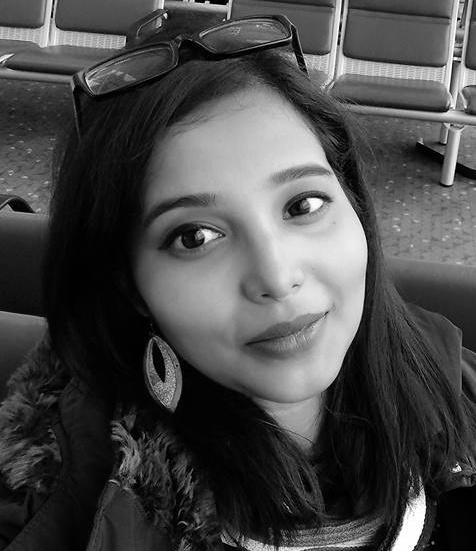}}]{Swatilekha Majumdar}
received the B.Tech degree in Electronics and Communication Engineering, the M.Tech degree in VLSI and Embedded Sytem from Indraprastha Institute of Information Technology (IIIT) Delhi, India, and the PhD degree in electronics from the Department of Electrical Engineering, Indian Institute of Technology (IIT) Delhi, India, in 2020. She is currently working as a Postdoctoral Researcher at IMEC, Belgium. Her research interests include exploiting emerging memory devices for digital, in-memory, machine learning, neuromorphic computing and computational neuroscience applications.
\end{IEEEbiography}

\begin{IEEEbiography}[{\includegraphics[width=1in,height=1.25in,clip,keepaspectratio]{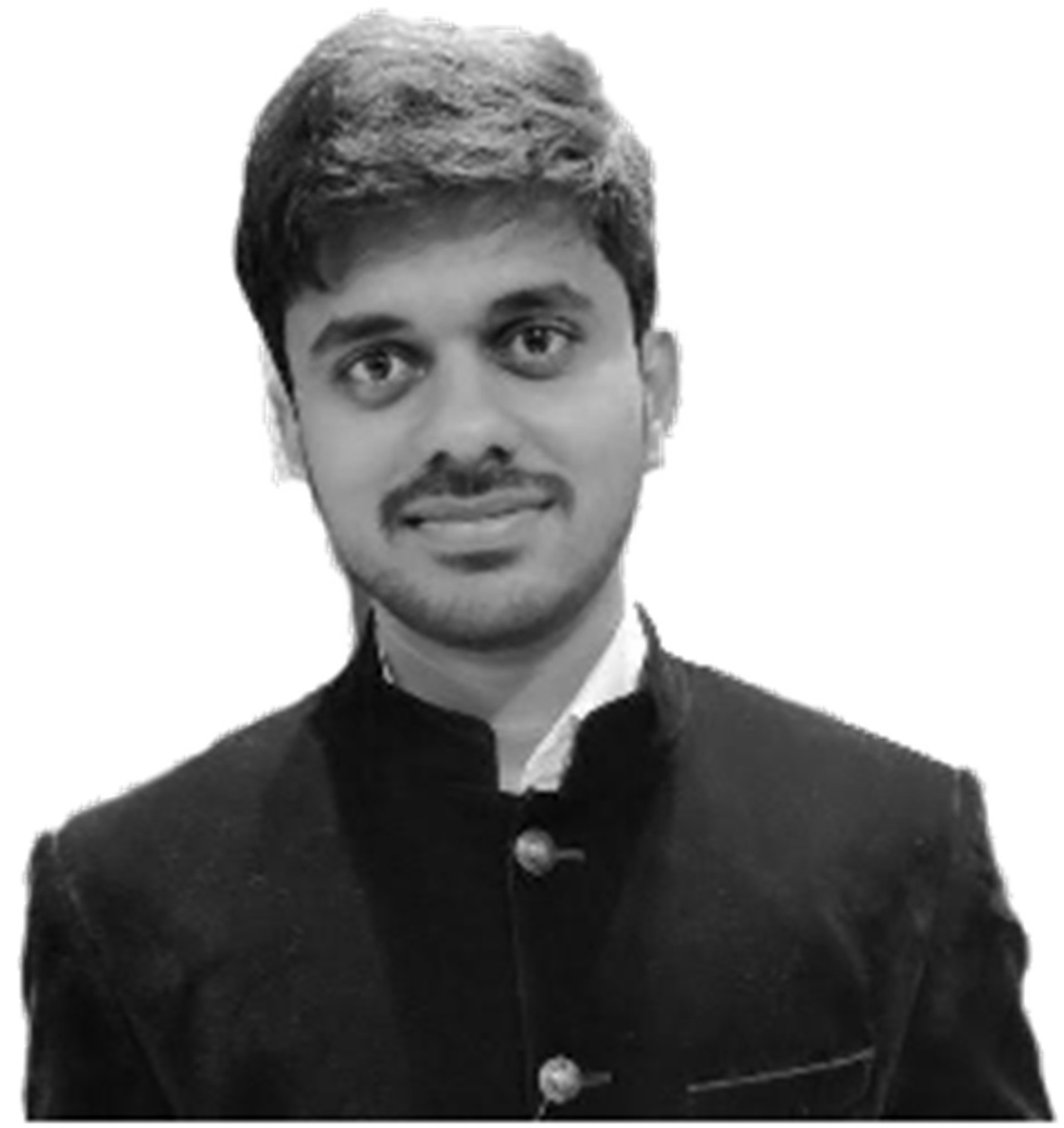}}]{Chetan Ralekar}
received B.Tech in Electronics and Telecommunication Engineering and M.Tech in Electronics Engineering. He is currently enrolled as a PhD student in the Department of Electrical Engineering, Indian Institute of Technology, Delhi, India. His research interests include artificial intelligence, pattern recognition, computational perception and cognition, neuroscience.
\end{IEEEbiography}

\begin{IEEEbiography}[{\includegraphics[width=1in,height=1.25in,clip,keepaspectratio]{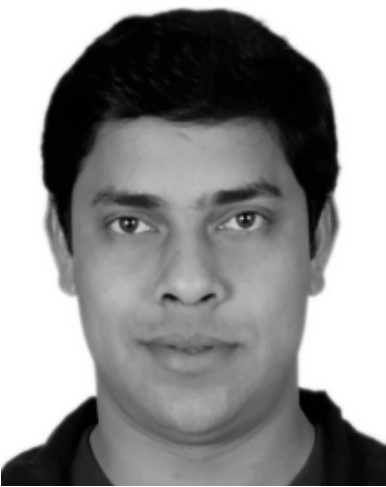}}]{Tapan Kumar Gandhi}
received the B.Sc. degree in physics, the M.Sc. degree in electronics, the M.Tech. degree in bioelectronics, and the Ph.D. degree in biomedical engineering from IIT Delhi, in 2001, 2003, 2006, and 2011, respectively. He is currently an Associate Professor with the Department of Electrical Engineering, IIT Delhi. Prior to joining IIT Delhi, as a Faculty Member, he was a Postdoctoral Fellow with the Massachusetts Institute of Technology (MIT), Cambridge, USA, for three years. His research interests include cognitive computation, artificial intelligence, medical instrumentation, biomedical signal and image processing, and assistive technology.

\end{IEEEbiography}


%\newpage



\vfill


\end{document}